\documentclass[10pt,twocolumn,letterpaper]{article}

\usepackage[pagenumbers]{cvpr} 

\usepackage[dvipsnames]{xcolor}
\usepackage{overpic}

\definecolor{cvprblue}{rgb}{0.21,0.49,0.74}
\usepackage[pagebackref,breaklinks,colorlinks,citecolor=cvprblue]{hyperref}

\newcommand{\figref}[1]{Fig.~\ref{#1}}
\newcommand{\tabref}[1]{Tab.~\ref{#1}}
\newcommand{\secref}[1]{Sec.~\ref{#1}}

\def\paperID{134} 
\def\confName{CVPR}
\def\confYear{2024}

\def\eg{\emph{e.g.,~}}
\def\ie{\emph{i.e.,~}}

\newcommand{\ourMthd}{TeMO}
\newcommand{\nameofmodule}{DGA}
\newcommand{\nameofloss}{CGC}

\newcommand{\linear}[1]{\mathrm{Linear}(#1)}
\newcommand{\myPara}[1]{\vspace{5pt}\noindent\textbf{#1}}

\newcommand{\namefontcolor}[1]{\textcolor{black}{#1}}

\title{\ourMthd: Towards Text-Driven 3D Stylization for Multi-Object Meshes }

\author{\href{https://zhangxuying1004.github.io/}{\namefontcolor{Xuying Zhang}} 
    \quad \href{https://yinbow.github.io/}{\namefontcolor{Bo-Wen Yin}} 
    \quad \href{http://www.fishworld.site/}{\namefontcolor{Yuming Chen}}  
    \quad \href{https://www.lin-zheng.com/}{\namefontcolor{Zheng Lin}}  
    \quad \href{https://mmcheng.net/}{\namefontcolor{Yunheng Li}} \\
    \quad \href{https://houqb.github.io/}{\namefontcolor{Qibin Hou}}\thanks{Qibin Hou is the corresponding author} 
    \quad \href{https://mmcheng.net/cmm/}{\namefontcolor{Ming-Ming Cheng}}   \\ 
    VCIP, School of Computer Science, Nankai University \\
}

\begin{document}
\maketitle
\begin{abstract}
\vspace{-2pt}
Recent progress in the text-driven 3D stylization of a single object has been considerably promoted by CLIP-based methods.
However, the stylization of multi-object 3D scenes is still impeded in that 
the image-text pairs used for pre-training CLIP mostly consist of an object.
Meanwhile, the local details of multiple objects may be susceptible to omission due to the existing supervision manner primarily relying on coarse-grained contrast of image-text pairs.
To overcome these challenges, we present a novel framework, 
dubbed \textbf{\ourMthd}, 
to parse multi-object 3D scenes and edit their 
styles under the contrast supervision at multiple levels. 
We first propose a Decoupled Graph Attention (\textbf{\nameofmodule{}}) module 
to distinguishably reinforce the features of 3D surface points.
Particularly, a cross-modal graph is constructed to align the object points accurately  
and noun phrases decoupled from the 3D mesh and textual description.
Then, we develop a Cross-Grained Contrast (\textbf{\nameofloss}) supervision system, 
where a fine-grained loss between the words in the textual description and 
the randomly rendered images are constructed to complement the coarse-grained loss.
Extensive experiments show that our method can synthesize high-quality stylized content and outperform the existing methods over a wide range of multi-object 3D meshes.
Our code and results will be made publicly available.
\end{abstract}
\vspace{-8pt}    

\section{Introduction}
\label{sec:intro}

3D asset creation through stylization aims to synthesize stylized content on the bare meshes 
to conform to the given text descriptions~\cite{michel2022text2mesh,lei2022tango}, 
referring images~\cite{wang2022clip,zhang2022arf}, or 3D shapes~\cite{yin20213dstylenet}.
This research plays an important role in a wide spectrum of applications, 
\eg virtual/augmented reality~\cite{chen2022instant,caetano2022arfy}, 
gaming industries~\cite{zhang2022novel}, and robotics~\cite{han2022submillimeter}.
Moreover, it also presents considerable potential and has attracted increasing attention 
in computer vision and graphics communities.
Considering the ready availability and expressiveness of text prompts 
as well as the popularity of large-scale Contrastive Language-Image Pre-training 
(CLIP)~\cite{radford2021learning} model, 
we choose to work with text-driven 3D stylization.
%

\newcommand{\wbText}[1]{\color{white}{\textbf{#1}}}

\begin{figure}[t]
  \centering
  \small
  \begin{overpic}[width=\linewidth]{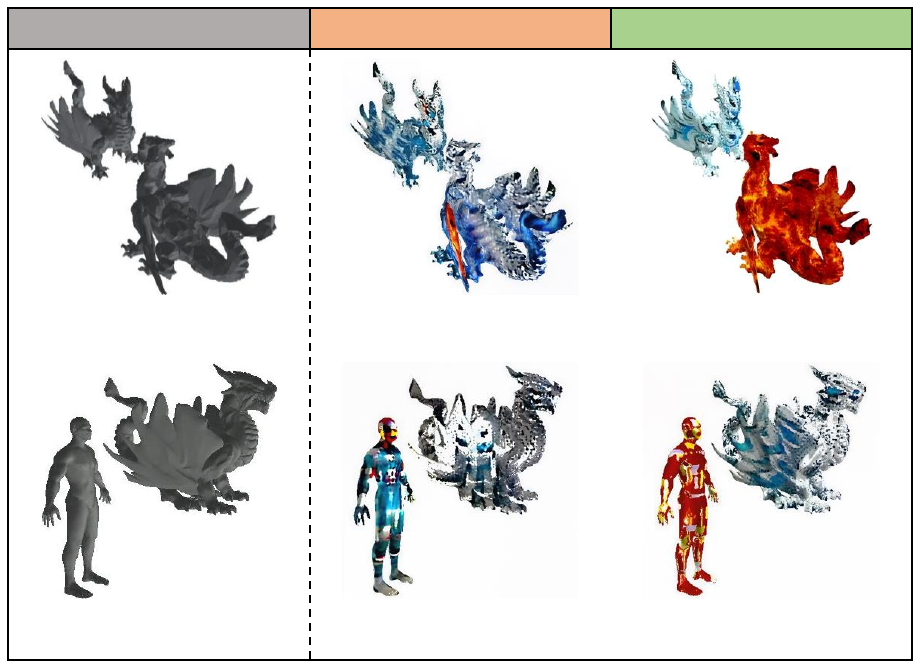}
    \put (8,68.4){\wbText{Bare Mesh}}
    \put (43,68.4){\wbText{TANGO}}
    \put (74,68.4){\wbText{Our \ourMthd}}
    \put (8,36){two dragons}
    \put (43,36){``a fire dragon and an ice dragon"}
    \put (5,3){person $\&$ dragon}
    \put (43,3){``an iron man and an ice dragon''}
    \end{overpic}
    \caption{Visual comparisons between the existing 3D stylization methods 
    (\eg TANGO~\cite{lei2022tango}) and our \ourMthd{} in multi-object scenes. 
    For a scene with multiple objects of the same/different categories, 
    existing methods are prone to interference between different properties of the objects, 
    while our \ourMthd{} is able to accurately synthesize the desired stylized 
    content for each object.
  }\label{fig:intro}
  \vspace{-10pt}
\end{figure}
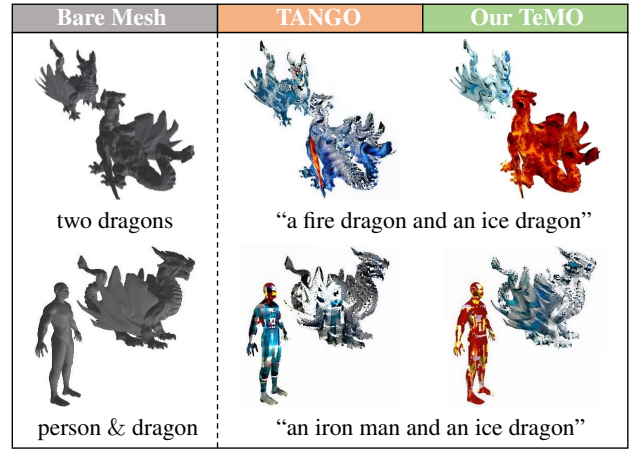

Recent years have witnessed the emergence of a series of impressive works
\cite{michel2022text2mesh,lei2022tango,ma2023x,chen2023fantasia3d}, 
aiming to drive the advancement of text-driven 3D stylization.
Existing methods usually adopt multi-layer perceptrons (MLPs) 
to predict the location attribute displacements of the bare mesh 
under the supervision of the contrastive loss in CLIP.
We observe that these works focus on the stylization of a single 3D object and 
perform poorly on multiple objects, 
as shown in the second column of \figref{fig:intro}.
We argue that two inherent characteristics of CLIP result in this issue: 
\textbf{i)} CLIP is mainly pre-trained with image-text pairs mostly consisting of a single object;
\textbf{ii)} CLIP loss employs global representation vectors from 
images and text to coarsely match these two modalities, 
which inevitably causes the loss of local details.
Moreover, the key to synthesizing desired styles for multiple 3D objects lies in the parsing of such 3D scenes and the multi-grained supervision for details refinement.

To simultaneously generate stylized content for multiple 3D objects, the primary step is to achieve accurate alignment between the objects in the 3D mesh and the target text. 
However, existing methods employ global semantics of the text to stylize a single object,
which inevitably produces noises when stylizing the objects in multi-object scenes.
To overcome this challenge, we propose to parse the 3D scene by introducing a Decoupled  Graph Attention (\nameofmodule{}) module.
Specifically, all noun phrases are decoupled from the text prompt 
and the mesh surface points of the current view are divided 
into several clusters as well.
Then, a cross-modal graph is constructed to establish 
connections of the noun phrases to their corresponding object points 
while distancing them from the irrelevant ones.
This graph enables the accurate interaction between two interrelated modalities. 
Finally, the surface point features of 3D objects can be reinforced by independent cross-attention fusions with their neighboring word nodes in the graph architecture.

Furthermore, we also design a Cross-Grained Contrast (\nameofloss{}) loss to perform comprehensive cross-modal supervision for the stylization of multiple objects.
The goal is to guide the network to generate 
more stylization details for multiple 3D objects to match the target text.
Our loss consists of two parts, 
\ie coarse-grained contrast and fine-grained contrast.
In the former part, the text prompt is regarded as sentence-level supervision, 
which calculates the similarity between the 2D views 
rendered from the stylized 3D mesh 
and the text prompt using the global feature vectors from the CLIP model.
In the latter part, we see the text prompt from the word level,
and consider the similarities between each word of the sentence 
and the rendered images of the view sets.
To be specific, we produce the word representations of the text prompt 
by taking the hidden states from the text encoder of CLIP.
Motivated by the recent process in video-text retrieval~\cite{ma2022x}, 
we calculate fine-grained loss via the weighted summation of the element 
in similarity vectors based on the importance of each word or image.

%
Based on the well-designed DGA module and CGC loss,
we propose a novel framework towards \textbf{Te}xt-Driven 3D stylization for \textbf{M}ulti-\textbf{O}bject Meshes, called \ourMthd.
To validate the effectiveness of our \ourMthd, 
extensive experiments are conducted on various multi-object 3D scenes, 
as shown in the $3^{rd}$ column of \figref{fig:intro}.
The experimental results demonstrate our \ourMthd{} is less susceptible 
to interference from multiple objects and can generate 
superior stylized assets compared with the existing 3D stylization methods.

Our contributions can be summarized as follows:
\begin{itemize}

\item We present a new 3D stylization framework, called \ourMthd. 
To the best of our knowledge, it is the first attempt to parse the objects in the text and 3D meshes and generate stylizations for multi-object scenes.

\item We propose a Decoupled Graph Attention (\nameofmodule) module, 
which constructs a graph structure to align the surface points 
in the multi-object mesh and the noun phrases in the text prompt. 

\item We design a Cross-Grained Contrast (\nameofloss) loss, in which the text is contrasted with the rendered images from sentence and word levels.

\end{itemize}

\section{Related Work}
\label{sec:related_work}

\subsection{Text-Driven 3D Manipulation}

Generating or editing 3D content according to a given prompt 
is a long-standing objective in computer vision and graphics.
Among all forms of the prompt, the text has garnered the most conspicuous gaze
due to three reasons:
i) Text descriptions are readily accessible from the existing corpus;
ii) Text descriptions are particularly user-friendly since they are easily modifiable 
and can effectively express complex concepts related to stylizations;
iii) The popularity of large-scale CLIP~\cite{radford2021learning} model 
has made achieving visual-language supervision possible.

Text2Mesh~\cite{michel2022text2mesh} proposes a neural-style field network 
to predict the color and displacement of mesh vertices.
TANGO~\cite{lei2022tango} proposes to disentangle the appearance style 
as the spatially varying bidirectional reflectance distribution, 
the local geometric variation, and the lighting condition.
Then, X-Mesh~\cite{ma2023x} integrates the target text guidance  
by utilizing text-relevant spatial and channel-wise attention 
during vertex feature extraction.
Motivated by the remarkable progress in text-driven 2D generation
\cite{ramesh2022hierarchical,rombach2022high}, 
TEXTure~\cite{richardson2023texture} and Text2Tex~\cite{chen2023text2tex} 
incorporate a pre-trained depth-aware image diffusion model 
to synthesize high-resolution partial textures 
from multiple viewpoints progressively.

To make full use of the priors in the pre-trained 2D text-to-image diffusion model, 
DreamFusion~\cite{poole2022dreamfusion} introduces a 
Score Distillation Sample (SDS) loss to perform text-to-3D synthesis.
With the help of SDS loss, 
Latent-NeRF~\cite{metzer2023latent} and Fantasia3D~\cite{chen2023fantasia3d} 
can generate 3D shapes and appearances for 3D objects.
Despite achieving impressive results, 
these methods focus on the stylization of a single 3D object 
and rarely explore multi-object scenes.
CLIP-Mesh~\cite{mohammad2022clip} attempts to generate multiple 3D objects for target text. 
Nevertheless, the resulting content is not satisfactory. 
In this paper, we parse the objects described in rendered images and text prompts, 
aligned by two well-designed strategies.

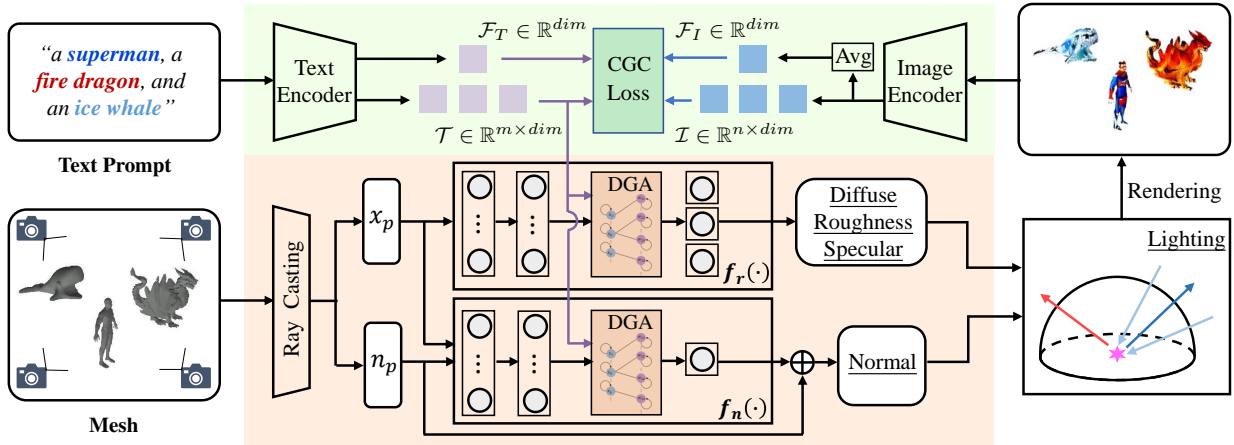
\begin{figure*}[t]
  \centering
  \footnotesize
  \begin{overpic}[width=0.94\linewidth]{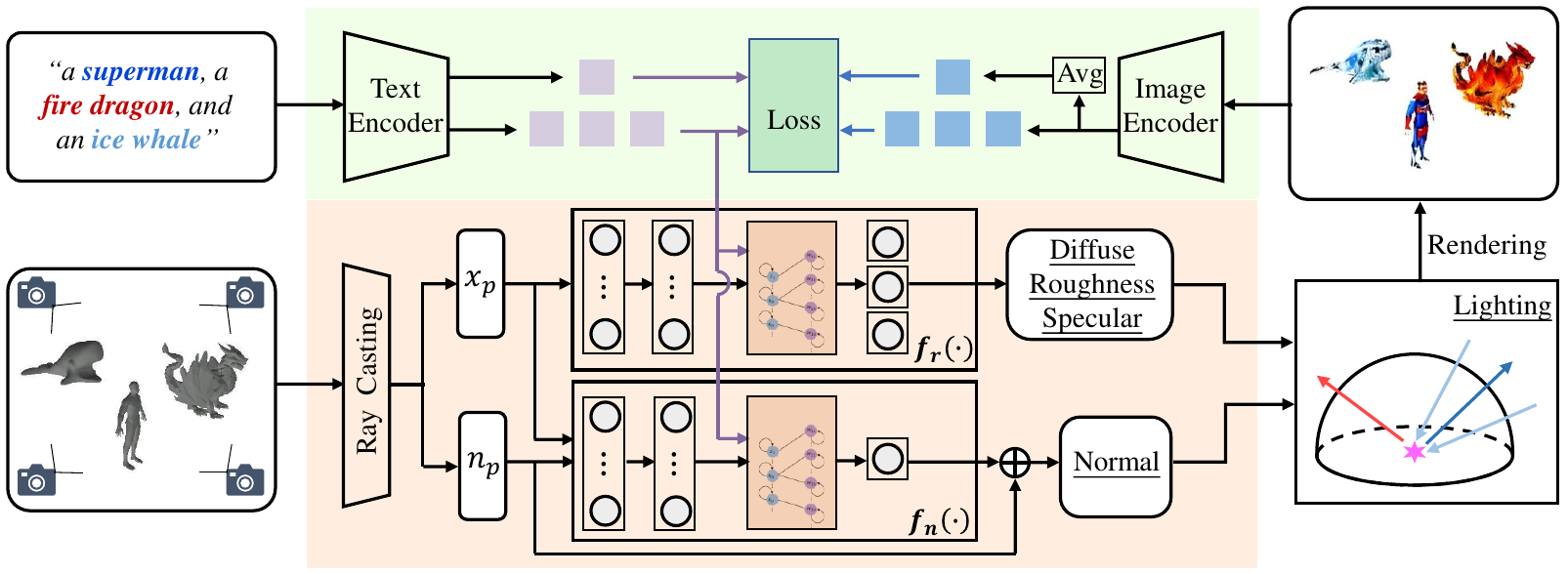}
    \put (7,1.5){\color{black}{\textbf{Mesh}}}
    \put (4.5,22.5){\color{black}{\textbf{Text Prompt}}}
    \put (35,25){$\mathcal{T} \in \mathbb{R}^{m \times dim}$}
    \put (54.5,25){$\mathcal{I} \in \mathbb{R}^{n \times dim}$}
    \put (38.5,33.5){$\mathcal{F}_{T} \in \mathbb{R}^{dim}$}
    \put (54.5,33.5){$\mathcal{F}_{I} \in \mathbb{R}^{dim}$}
    \put (49,30.7){\nameofloss{}}
    \put (49,21){\nameofmodule{}}
    \put (49,9.8){\nameofmodule{}}
  \end{overpic}
  \vspace{-5pt}
  \caption{The overall architecture of the proposed \ourMthd{} framework. 
    We first specify several cameras to cast rays toward the objects in the 3D mesh scene. 
    Then, a surface point $x_{p}$ and normal $n_{p}$ can be attained 
    from each ray intersected with the objects.
    These points and normals are fed to the attribute prediction network 
    where the features of 3D objects are parsed and 
    interacted with the decoupled text features via our proposed \nameofmodule{} module.
    Meanwhile, we employ a series of spherical Gaussians to represent the lighting. 
    Finally, a differentiable SG render is adopted to render images, 
    which are utilized to contrast with the text prompt by our designed \nameofloss{} 
    loss.
  }\label{fig:overall}
  \vspace{-5pt}
\end{figure*}

\subsection{Attention Mechanism}

The concept of the attention mechanism was initially introduced in 
neural machine translation~\cite{bahdanau2014neural}, 
where the weighted summation of the candidate vectors is calculated 
according to their importance scores.
This technology has been extended to a myriad of tasks, 
\eg natural language processing~\cite{luong2015effective,vaswani2017attention,devlin2018bert}, 
computer vision~\cite{hu2018squeeze,liu2021swin,yin2022camoformer,zhang2023referring}, 
and multi-modal learning~\cite{luo2021dual,zhang2021rstnet,yin2023dformer,wu2022difnet}.
For instance, Transformer~\cite{vaswani2017attention} employs the self-attention operation 
to establish connections between words within a sentence and 
utilize the cross-attention mechanism to align source and target sentences.
%
%
Non-local network~\cite{wang2018non} takes the lead in introducing self-attention 
to computer vision and achieves great success in video understanding and object detection.
ViT~\cite{dosovitskiy2020image} treats an image as a sequence of patches and 
employs a Transformer encoder based on self-attention to perform image classification.
Swin Transformer~\cite{liu2021swin} introduces shifted windows to 
enhance the local perception ability of self-attention.
More recently, X-Mesh~\cite{ma2023x} designs 
a text-guided dynamic attention mechanism for vertex feature extraction of a 3D object.
However, this guidance only relies on a text feature vector 
without considering the parsing of text and 3D scenes. 
In this paper, the multiple objects decoupled from the target text and 3D mesh 
are aligned via a cross-modal graph to achieve precise guidance.

\subsection{Multi-modal Contrastive Learning}

Contrastive learning has become an increasingly popular research topic in 
the multi-modal community due to its ability to align different modal representations.
Based on this strategy, CLIP~\cite{radford2021learning} is pre-trained on 
an abundance of image-text pairs, achieving great success in cross-modal supervision.
TACo~\cite{yang2021taco} presents a token-aware cascade contrastive learning 
based on the syntactic classes of words to achieve fine-grained semantic alignment 
in text-video retrieval.
Concurrently, FILIP~\cite{yao2021filip} proposes comparing the image patches 
with the words in the sentence. 
Regarding the text-driven 3D stylization, the CLIP loss, 
which calculates the similarity between the image and text vectors in the embedding space of CLIP, 
is adopted by the vast majority of methods.
Although achieving impressive results in stylizing a single object, 
these methods cannot be well adapted to scenes with multiple 3D objects.
We argue an important reason for this issue is the loss of local details 
caused by such coarse-grained supervision.
In this paper, we propose a cross-grained supervision strategy, 
which considers fine-grained and coarse-grained contrasts 
to achieve a more precise semantic alignment between rendered image and text.
\section{Methodology}

\subsection{Overall Architecture} \label{subsec:overall}

\figref{fig:overall} shows the end-to-end architecture of our \ourMthd{} framework.
Given a bare mesh and a text prompt containing multiple objects, the \ourMthd{} aims to synthesize stylization on the mesh to match the text descriptions.
We employ a set of vertices $V \in \mathbb{R}^{e \times 3}$ and faces $F \in \{ 1,...,e \}^{u \times 1}$ to explicitly define the input triangle mesh, which is fixed throughout the training.
Following TANGO~\cite{lei2022tango}, we disentangle the appearance style as the spatially varying bidirectional reflectance distribution function~\cite{boss2021nerd,zhang2021nerfactor,zhang2021physg} (including diffuse, roughness, specular), the local geometric variation (normal map), and the lighting condition.

\begin{figure*}[t]
    \centering
    \includegraphics[width=.98\linewidth]{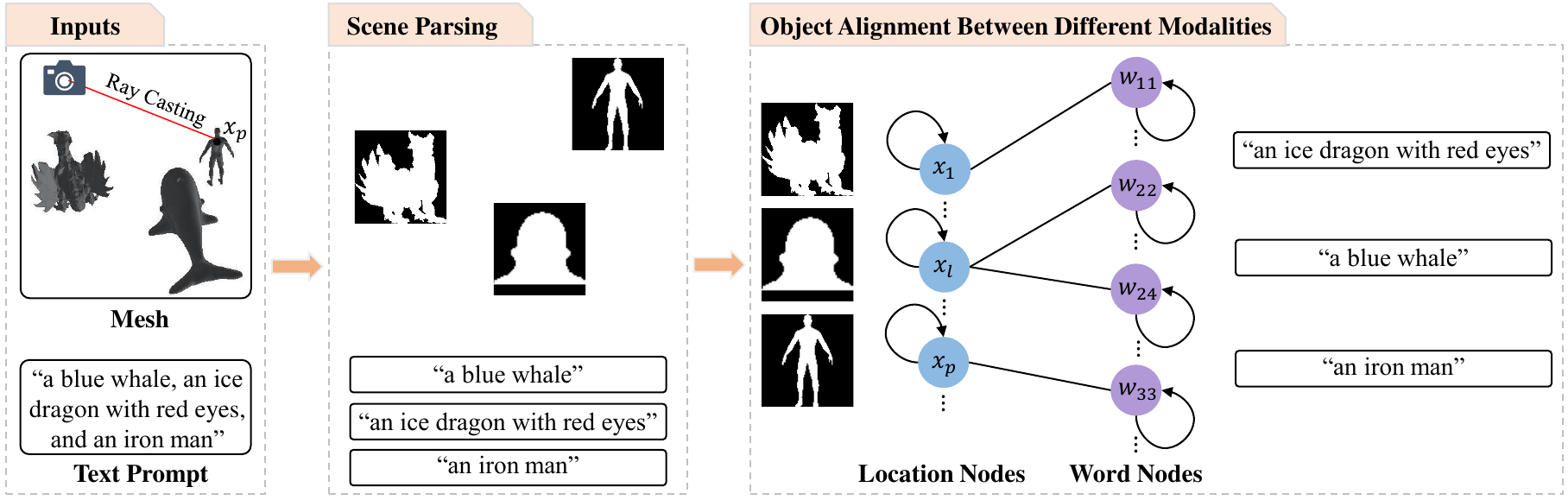}
    \vspace{-5pt}
    \caption{
    Construction pipeline of the cross-modal graph architecture in our \nameofmodule{} module.
    Note that $x_{p}$, the surface point of the 3D objects, and $w_{ij}$, the $j$-th word in the $i$-th noun phrase, are connected together only if they correspond to the same object.
    } 
  \label{fig:dgca}
  \vspace{-5pt}
\end{figure*}

We start by normalizing the vertex coordinates to lie inside a unit sphere.
Then, we randomly sample points around the mesh using Gaussian distribution as camera positions to render images. 
Next, we can obtain a camera ray $R_{p} = \{ c + t\nu_{p} \}$ from the sampled camera position $c$ and a pixel $p$ in rendered images, where $\nu_{p}$ is the direction of the ray.
Further, ray casting~\cite{roth1982ray} is used to seek out the ray and mesh's first intersection point and intersection face.
Moreover, the normal $n_{p} \in \mathbb{R}^{3}$ of the intersection face is employed as surface normal at the point $x_{p} \in \mathbb{R}^{3}$.

To achieve multiview-consistent features, our \ourMthd{} is restricted to predicting the normal displacement as a function of the location, while allowing the color materials to be predicted as a function of both location and viewing direction.
Therefore, our \ourMthd{} represented as MLPs includes two branches, \ie normal branch $f_{n}(\cdot)$ and reflectance branch $f_{r}(\cdot)$.
Specifically, the former is utilized as the prediction of normal offset on the point $x_{p}$, and the latter is designed to predict surface reflectance coefficients of the material at the location $x_{p}$, \ie diffuse, roughness, and specular.
To synthesize high-frequency details, we also apply the Fourier positional encoding~\cite{tancik2020fourier} to every input. 
In addition, the spherical Gaussian is employed to represent each light intensity $L_{i}(\cdot)$ due to its closed-form nature and analytical solution.
Based on the attained geometric and appearance components, each pixel color in the rendered image can be calculated by a hemisphere renderer~\cite{lei2022tango}:
\begin{align}
   \!L_{p}(\nu_{p},x_{p},n_{p}) &=\!\int_{\Omega}\!L_{i}(w_{i})f_{r}(\nu_{p},\!w_{i},\!x_{p})(w_{i}\!\cdot \!\hat{n}_{p})dw_{i}, \\
   \hat{n}_{p} &= n_{p} + f_{n}(x_{p}, n_{p}),
  \label{eq:render}  
\end{align} 
where $\Omega = \{ w_{i}: w_{i} \cdot \hat{n}_{p} \geq 0 \}$ represents the hemisphere, $w_{i}$ is the incident light direction, and $\hat{n}_{p}$ is the estimated normal on surface point $x_{p}$.

\subsection{Decoupled Graph Attention} \label{subsec:module}
To achieve text-drive stylization for multiple 3D objects, the key issue that needs to be solved is the accurate alignment between the objects described in the text and those in the meshes.
X-Mesh~\cite{ma2023x} has incorporated text-guided dynamic linear layers, in which the global representation vector of the target object in the text is utilized as guidance to acquire text-aware vertex features.
Nevertheless, the global vector containing information about multiple objects is prone to mutual interference and produces semantic noises during guidance for multi-object scenes.

To address this challenge, we propose to parse the objects in the text and mesh.
We first extract the noun phrases modified by adjectives or prepositional phrases from the text using the NLTK tools~\cite{bird2009natural}.
Then, we employ the Gaussian Mixture Model (GMM)~\cite{zivkovic2004improved} to cluster the intersection point set $\{ x_{1}, ..., x_{p}, ... \}$ of the current ray and the mesh.
Meanwhile, we can obtain a binary map of objects in the current view based on whether the ray intersects with the mesh.
Further, we can decouple the objects in the binary map according to the clustered points and acquire several binary maps of individual objects.
Based on the disentangled noun phrases and binary maps for multiple objects, we can match the correct pairs by their semantic similarities.
As a result, the objects described in the text are aligned with their corresponding objects in the mesh, which are utilized to construct a cross-modal graph $G = (\mathcal{V}, \mathcal{E})$, as shown in
\figref{fig:dgca}.
To be specific, all surface point features and word features are considered as independent nodes to form the node set $\mathcal{V}$.
For the edge set $\mathcal{V}$, the link between the surface point node and the word node will be built if the semantic objects they belong to are the same.

Under the setting of this cross-modal graph, we can individually perform cross-attention between the surface point nodes and their neighboring word nodes, where the parsed surface point features are used as queries, and the parsed text features serve as keys and values.
The enhancement of the surface point node $v_{i} \in \mathbb{R}^{dim}$ can be formulated as:
\begin{align}
    \hat{v}_{i} &= \sum_{v_{j} \in \mathrm{Adj}(v_{i})} \alpha_{ij}\linear{v_{j}},\\
    \alpha_{ij} &= \frac{e^{\mathcal{W}_{ij}}}{\sum_{v_{j} \in \mathrm{Adj}(v_{i})}e^{\mathcal{W}_{ij}}}, \\
    \mathcal{W}_{ij} &= \frac{\linear{v_{i}}\linear{v_{j}}^\mathrm{T}}{\sqrt{d_{l}}},
    \label{eq:dgca}  
\end{align}
where $\mathrm{Adj}(v_{i})$ is the adjacency nodes of $v_{i}$ and $\mathrm{Linear}(\cdot)$ represents a linear
transformation.
With this attention mechanism, the surface point features of different objects in the mesh can be distinguishably reinforced under the guidance of the word features in the parsed text.

\subsection{Cross-Grained Contrast Supervision} \label{subsec:loss}
To guide the optimization of the neural network for 3D stylization, the first step is to render the stylized 3D mesh from multiple 2D views.
Most existing methods usually employ the visual encoder and text encoder of CLIP~\cite{radford2021learning} to extract global feature vectors for the rendered image and target text, respectively, which are contrasted to perform cross-modal supervision via cosine similarity: 
\begin{equation}
   \mathcal{L}_{coarse} = - \frac{\mathcal{F}_{I} \cdot \mathcal{F}_{T}}{\Vert \mathcal{F}_{I} \Vert_2 \Vert \mathcal{F}_{T} \Vert_2},
  \label{eq:coarse_loss}  
\end{equation} 
where $\mathcal{F}_{I} \in \mathbb{R}^{512}$ is the averaged feature vector of the images rendered from different views, $\mathcal{F}_{T} \in \mathbb{R}^{512}$ denotes the global feature vector of the target text, and $\Vert \cdot \Vert_2$ represents the Euclidean norm function. 

Although achieving impressive results for stylizing a single 3D object, these methods still have limitations in multi-object scenes.
Considering that a single feature vector still represents a sentence describing multiple objects, the object details may be lost in large amounts.
Therefore, such a coarse-grained contrast supervision is insufficient to guide the neural  network in synthesizing photorealistic stylized content for multiple 3D objects.

To solve this issue, we construct a fine-grained contrast supervision to complement the coarse-grained one.
%
%
Specifically, we first calculate the correlation map, \ie $\mathcal{S} \in \mathbb{R}^{n \times m}$ between the word features in the text and the visual features of the rendered images, which are also extracted from the text encoder and visual encoder of the CLIP:
%
\begin{align}
   \mathcal{S} = \frac{\mathcal{I} \cdot \mathcal{T}^\mathrm{T}}{\Vert \mathcal{I} \Vert_2 \Vert \mathcal{T} \Vert_2},
  \label{eq:fine_loss_sim}  
\end{align}
where $\mathcal{I} \in \mathbb{R}^{n \times 512}$ represents the features of the images rendered from $n$ views, $\mathcal{T} \in \mathbb{R}^{m \times 512}$ indicates the features of $m$ words in the text.
Then, we normalize the correlation matrix along the image axis and the text axis, respectively, to retrieve the text of interest and visual components.
This process can be formulated as:
\begin{align}
    \mathcal{S}_{I}(i) &= \frac{\sum_{k=1}^{m} \mathcal{S}(i, k)}{m}, \\
   \mathcal{S}_{T}(j) &= \frac{\sum_{k=1}^{n} \mathcal{S}(k, j)}{n}.
  \label{eq:fine_loss_sim1}  
\end{align}
Inspired by~\cite{ma2022x}, we further calculate an image-centered fine-grained contrast score and a text-centered fine-grained contrast score by the weighted summation of the similarity vectors, which can be formulated as follows:
\begin{align}
    \mathcal{L}_{I} &= \sum_{i=1}^{n} \frac{e^{\mathcal{S}_{I}(i)}}{\sum_{k=1}^{n} e^{\mathcal{S}_{I}(k)}}\mathcal{S}_{I}(i), \\
   \mathcal{L}_{T} &= \sum_{j=1}^{m} \frac{e^{\mathcal{S}_{T}(j)}}{\sum_{k=1}^{m} e^{\mathcal{S}_{T}(k)}}\mathcal{S}_{T}(j),
  \label{eq:fine_loss_score}  
\end{align}
where the weights are defined as the degree of correlation between the central and another modality.
Finally, we adopt the average value of these two scores as the fine-grained contrast loss, which can be defined as:
\begin{align}
     \mathcal{L}_{fine} = - (\mathcal{L}_{I} + \mathcal{L}_{T}) / 2.
  \label{eq:fine_loss}  
\end{align}

The coarse-grained and fine-grained contrast supervision complement each other to build a cross-grained contrast supervision system.
The former is utilized to align the global semantic information of the target text with the 3D objects, and the latter is used to achieve the local semantic alignment.
This loss can be defined as:
\begin{equation}
   \mathcal{L}_{cgcs} = \lambda_{c} \mathcal{L}_{coarse} + \lambda_{f} \mathcal{L}_{fine}, 
  \label{eq:cgcs}  
\end{equation} 
where $\lambda_{c}$ and $\lambda_{f}$ are two hyper-parameters to balance the cross-grained and the fine-grained losses, set to 1.0 and 0.33, respectively.
\section{Experiments}
\subsection{Experiment Setup}\label{subsec:setting}
\myPara{Datasets.}
To examine our method across a diverse set of 3D scenes, we first collect 3D object meshes from a variety of sources, \ie COSEG~\cite{sidi2011unsupervised}, Thingi10K~\cite{zhou2016thingi10k}, Shapenet~\cite{chang2015shapenet}, Turbo Squid~\cite{turbosquid2021turbosquid}, and ModelNet~\cite{wu20153d}.
Then, we randomly place several objects from the collected 3D set into a mesh using Blender. 
Note that we down-sample the number of meshes’ vertices and faces to ensure the robustness of our \ourMthd{} to low-quality meshes and reduce the burden of GPU during the stylization.
The meshes used in this paper contain an average of 79,303 faces, 16\% non-manifold edges, 0.2\% non-manifold vertices, and 12\% boundaries.

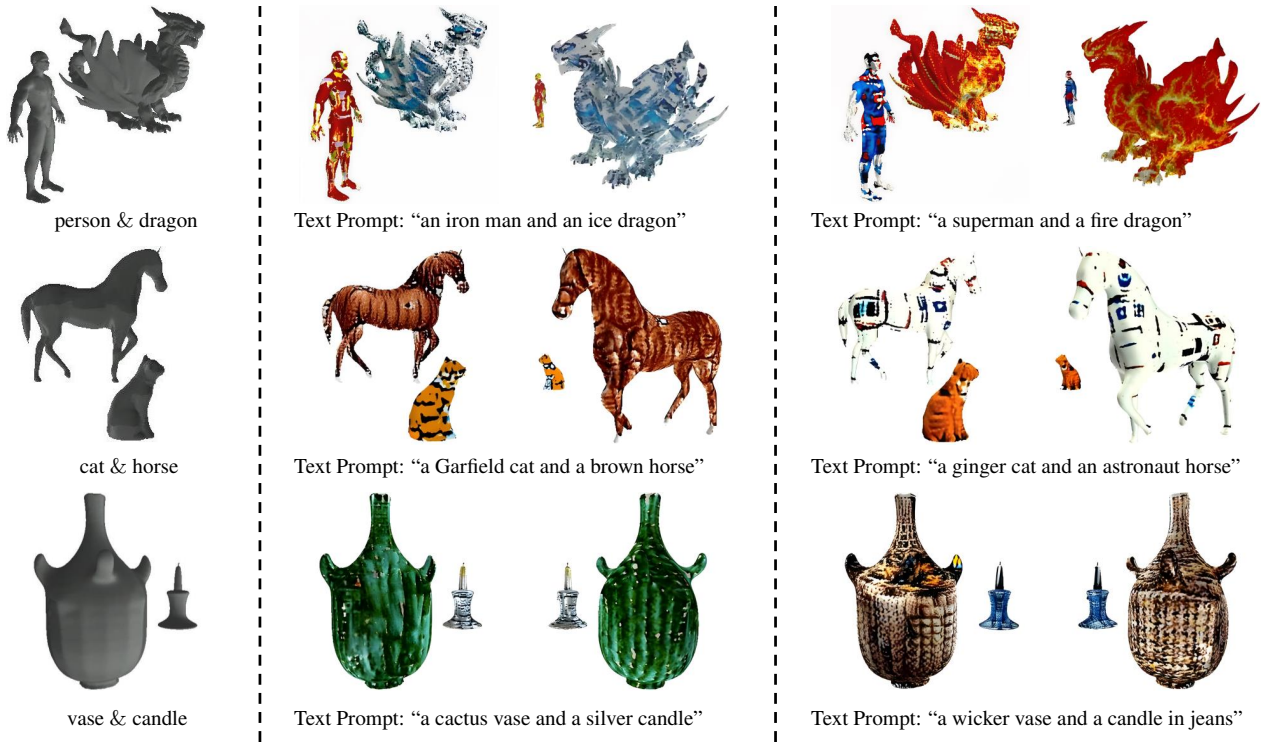
\begin{figure*}[t]
    \centering
    \footnotesize
    \begin{overpic}[width=0.96\linewidth]{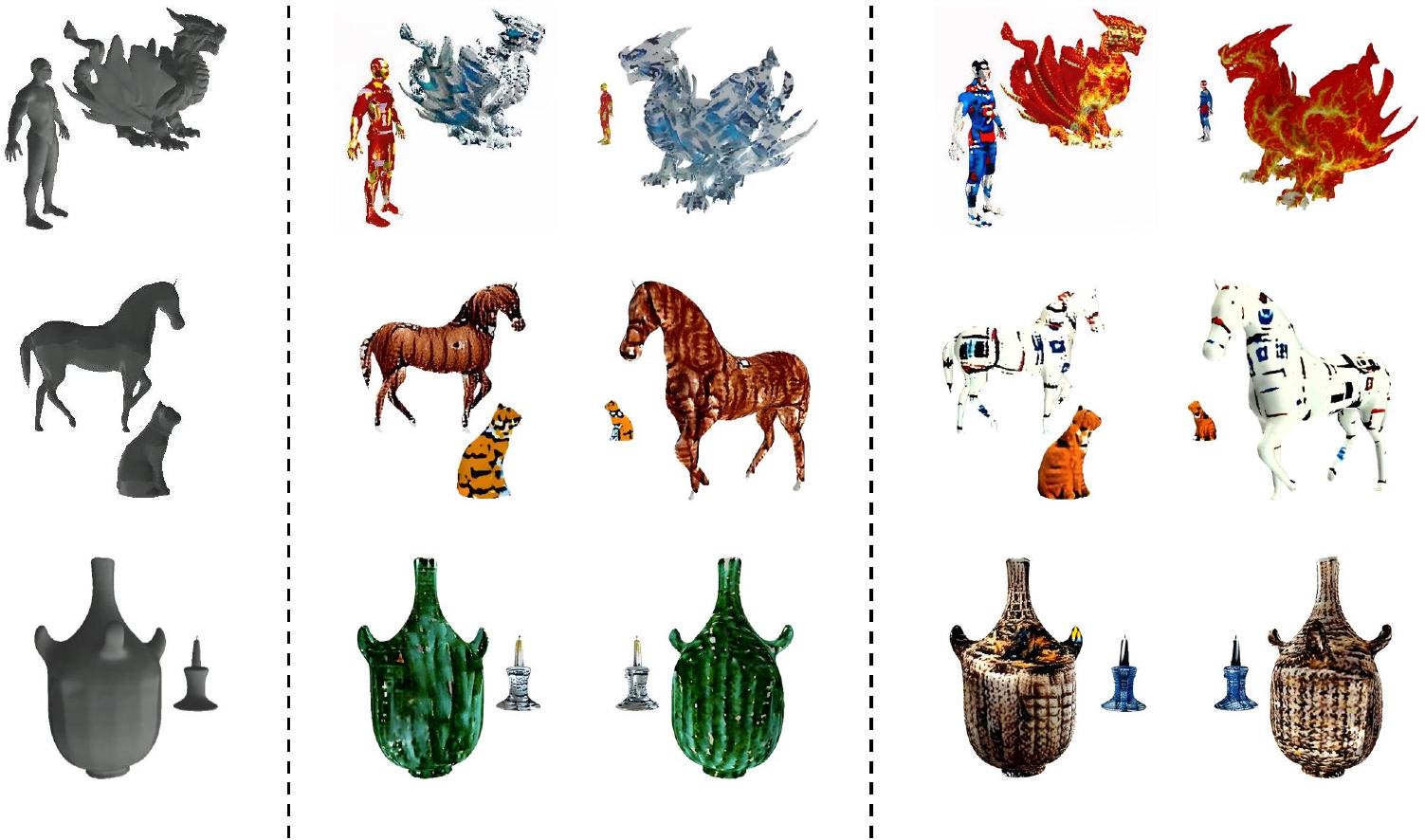}
    \put (5,1.5){vase $\&$ candle}
    \put (23,1.5){Text Prompt: ``a cactus vase and a silver candle''}
    \put (64,1.5){Text Prompt: ``a wicker vase and a candle in jeans''}
    \put (6,21.5){cat $\&$ horse}
    \put (23,21.5){Text Prompt: ``a Garfield cat and a brown horse''}
    \put (64,21.5){Text Prompt: ``a ginger cat and an astronaut horse''}
    \put (4,41){person $\&$ dragon}
    \put (23,41){Text Prompt: ``an iron man and an ice dragon''}
    \put (64,41){Text Prompt: ``a superman and a fire dragon''}
    \end{overpic}
    \caption{
    Given the same bare mesh, our \ourMthd{} produces stylized contents of various types for multi-object scenes to conform to the different text prompts.
    Please refer to \emph{supplementary materials} for a detailed version.
    } 
  \label{fig:disp}
\end{figure*}


\myPara{Implementation Details.}
Following the TANGO~\cite{lei2022tango} network, we adopt 3 linear layers with 256 dimensions to build the normal estimation branch. 
In the reflectance branch, the point features are extracted by 2 shared layers with 256 dimensions, followed by 3 exclusive layers to predict diffuse, specular, and roughness.
The dimension of our \nameofmodule{} module is also set as 256.
The word features in our \nameofmodule{} module are extracted from the text encoder of CLIP, and so are the ones in our \nameofloss{} loss.
We choose ViT-B/32 as the backbone of the pre-trained CLIP model in this paper, which is consistent with previous works~\cite{michel2022text2mesh,lei2022tango,ma2023x}.
We also process the rendered images with 2D augmentation strategies~\cite{frans2022clipdraw,lei2022tango} before feeding them into the pre-trained CLIP model.
Our \ourMthd{} model is optimized with the AdamW~\cite{loshchilov2017decoupled} strategy for 1500 iterations, where the learning rate is initialized to $5 \times 10^{-4}$ and decayed by 0.7 every 500 iterations. 
The entire training process takes about 10 minutes on a single NVIDIA RTX 3090 GPU.

\subsection{Qualitative Evaluation} \label{subsec:qual}
We conduct visualization experiments on a wide spectrum of multi-object scenes to verify the effectiveness of our \ourMthd{}.
However, we observe that the 3D symmetry prior used widely in previous works~\cite{michel2022text2mesh,lei2022tango} can cause interference between different parts during the stylization process of multiple objects.
We argue that the multiple objects of the meshes used in this paper are randomly placed to simulate a real 3D scene rather than along the z-axis.
To avoid this issue, we remove this prior in our \ourMthd{} and previous methods involved in the comparison.

\begin{figure*}[t]
    \centering
    \small
    \subfloat[Mesh: two dragons; Text Prompt: ``a fire dragon and an ice dragon''.]{
        \begin{overpic}[width=0.98\linewidth]{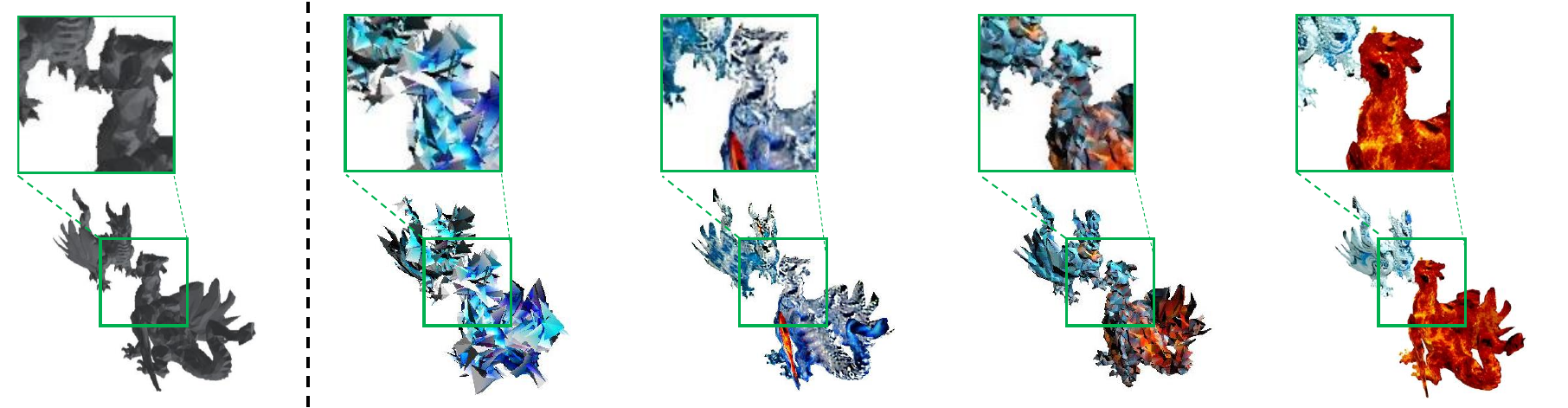}
        \put (2.4,26){\color{black}{\textbf{Bare Mesh}}}
        \put (21.6,26){\color{black}{\textbf{Text2Mesh}}~\cite{michel2022text2mesh}}
        \put (42.6,26){\color{black}{\textbf{TANGO}}~\cite{lei2022tango}}
        \put (63.5,26){\color{black}{\textbf{XMesh}}~\cite{ma2023x}}
        \put (83.2,26){\color{black}{\textbf{Our \ourMthd{}}}}
        \end{overpic}
    } \\
    \vspace{3pt}
    \subfloat[Mesh: vase $\&$ candle; Text Prompt: ``a wood vase and a brick candle''.]{
        \includegraphics[width=0.98\linewidth]{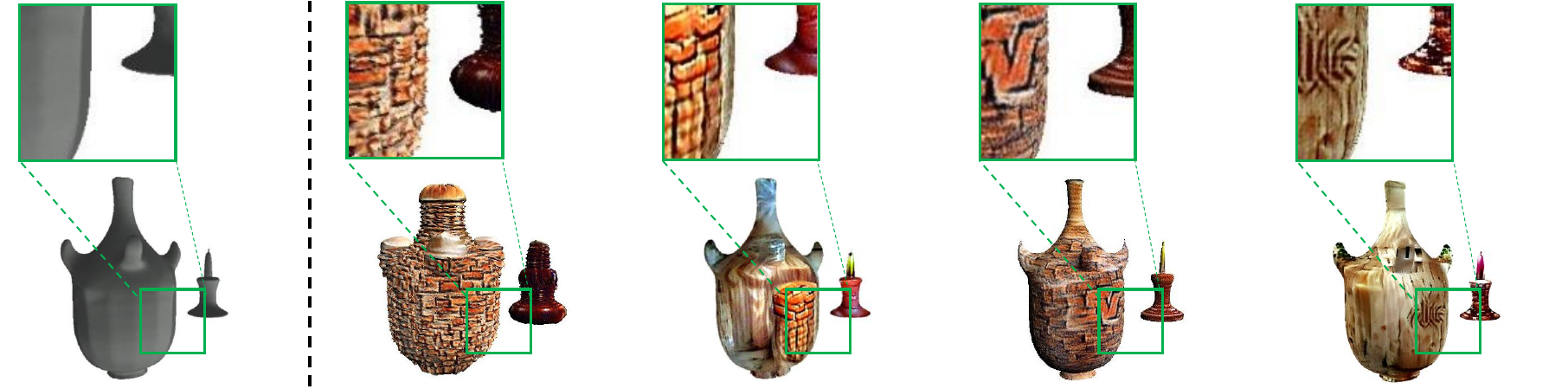}
    } \\
    \vspace{3pt}
    \subfloat[Mesh: person $\&$ dragon $\&$ whale; Text Prompt: ``a superman, a fire dragon, and an ice whale''.]{
        \includegraphics[width=0.98\linewidth]{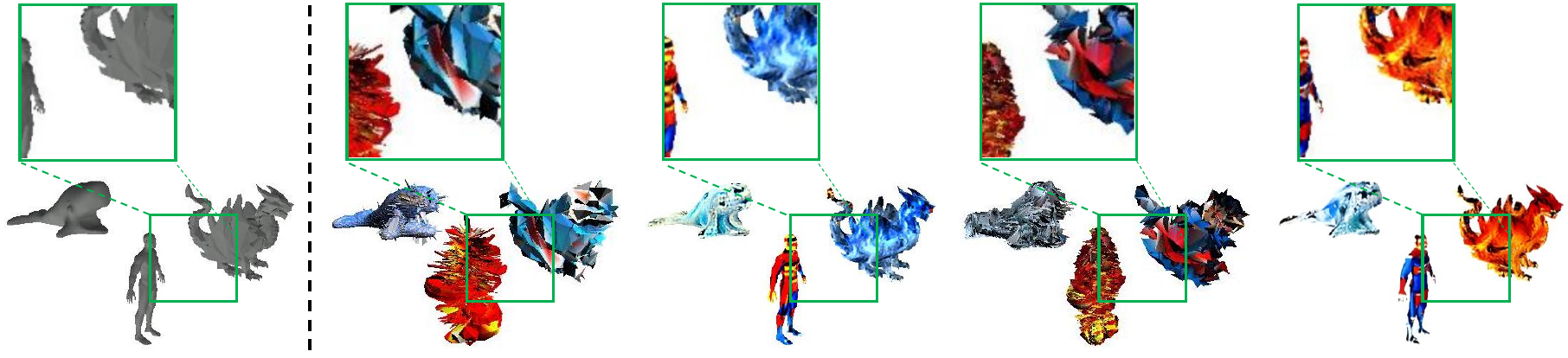}
    }
    \caption{Visual comparisons of our \ourMthd{} with previous text-driven 3D stylization methods on several multi-object scenes, including two objects of the same or different categories, and three different objects. 
    See \emph{supplementary materials} for more comparisons.
    } 
  \label{fig:comp1}
\end{figure*}

\myPara{Neural Stylization and Controls.}
We present the stylization results of our \ourMthd{} driven by different text prompts for the same multi-object mesh in \figref{fig:disp}.
As shown in the $1^{st}$ row where the 3D scene is composed of a person and a dragon, our \ourMthd{} can accurately distinguish between the person object and the dragon object and appropriately stylize different body parts for them according to the semantic roles described in each text prompt. 
Meanwhile, our \ourMthd{} also synthesizes desired stylizations for the 3D objects in the cat-horse mesh and vase-candle mesh, as shown in the $2^{nd}$ and $3^{rd}$ rows.
Moreover, \figref{fig:entire}, \figref{fig:disp_more}, and \figref{fig:details} of the supplementary material also show our \ourMthd{} stylizes the entire 3D scene and generates renderings with accurate details.
These experimental results demonstrate that our \ourMthd{} method can generate photorealistic details with fine granularity and maintain global semantic understanding for the given multi-object 3D scene.

\myPara{Qualitative Comparisons.}
We first provide the visual comparisons of prediction results between our \ourMthd{} and previous pioneering works in text-driven 3D stylization, including Text2Mesh~\cite{michel2022text2mesh}, TANGO~\cite{lei2022tango}, and X-Mesh~\cite{ma2023x}. 
To ensure a fair comparison, we adopt the official implementations of these methods and also train them with the default settings without the symmetry prior.
The experimental results show it is a real struggle for Text2Mesh~\cite{michel2022text2mesh} and TANGO~\cite{lei2022tango} to understand the detailed semantics of the text prompt with multiple objects.
As shown in the $1^{st}$ row of \figref{fig:comp1} where the 3D scene contains two objects of the same category, given a text prompt ``a fire dragon and an ice dragon", they tend to capture the ``ice" property, missing the ``fire" property.
For a 3D scene containing two objects of different categories, they are prone to mixing the properties of these objects, as shown in the $2^{nd}$ row where the text prompt is ``a wood vase and a brick candle".
Therefore, the stylized assets they generate for these multi-object scenes are unsatisfactory.
X-Mesh generates more accurate results that align with the text prompts, as shown in the $1^{st}$ and $2^{nd}$ rows, which can be attributed to incorporating the text vector while extracting vertex features. 
However, it can produce semantic noises due to its utilization of the text vector containing attributes of multiple objects to process all vertex features.
With an increasing number of objects, it will also encounter challenges related to comprehending text details and the alignment between the text and 3D objects, as shown in the $3^{rd}$ row.
Besides, we compare our \ourMthd{} with recent representative 3D stylization methods based on diffusion strategies~\cite{metzer2023latent,richardson2023texture,chen2023fantasia3d}, as shown in \figref{fig:comp2} of the supplementary materials.
These methods still fail to generate stylized assets without mixed properties, which can also be attributed to the global semantic guidance of the target text for the stylization of multiple 3D objects.
In contrast, our \ourMthd{} equipped with 3D scene parsing and multi-grained supervision, can generate photorealistic stylized content for each object in these 3D scenes to conform to the descriptions in the text prompts.
%

\subsection{Quantitative Evaluation} \label{subsec:quan}
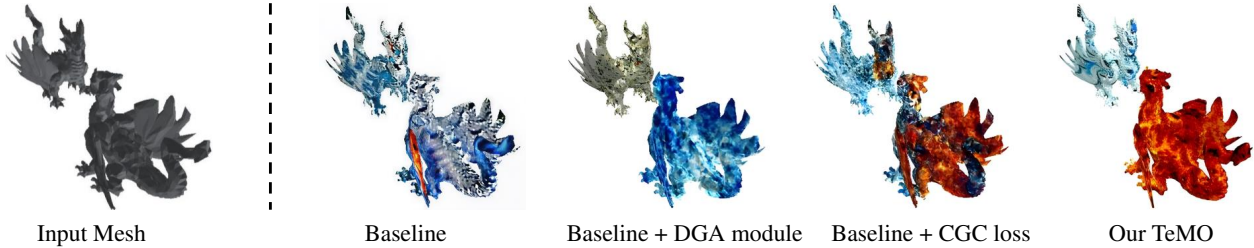
\begin{figure*}[t]
    \centering
    \small
    \begin{overpic}[width=0.96\linewidth]{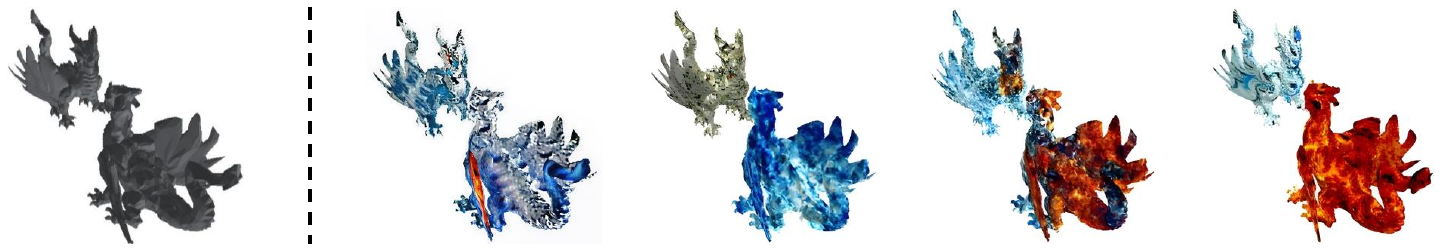}
    \put (3,-2){Input Mesh}
    \put (29,-2){Baseline}
    \put (45,-2){Baseline~+~\nameofmodule{} module}
    \put (66,-2){Baseline~+~\nameofloss{} loss}
    \put (88,-2){Our \ourMthd{}}
    \end{overpic}
    \vspace{13pt}
    \caption{Ablation experiments on the proposed designs of our \ourMthd{}. 
    Mesh: two dragons;
    Text Prompt: ``a fire dragon and an ice dragon''. } 
  \label{fig:abl}
\end{figure*}


\myPara{Objective Metric.}
We adopt the CLIP score to objectively evaluate the semantic alignment achieved by our \ourMthd{} and recent 3D stylization methods.
Specifically, 8 views spaced $45^{\circ}$ around the stylized meshes are chosen to obtain the rendered 2D images.
Then, the visual objects are compared with the textual objects in CLIP's embedding space via the cosine function.
As shown in the $2^{nd}$ column of \tabref{tab:quan}, our \ourMthd{} surpasses previous methods by a large margin.
These results demonstrate the superiority of our \ourMthd{} over existing methods on multi-object stylization.

\begin{table}[tp]
\setlength\tabcolsep{4.5pt}
\centering
\small
\caption{Quantitative comparisons of our \ourMthd{} and previous text-driven 3D stylization methods in multi-object scenes, including an objective alignment score (0-1) and three subjective opinion scores (1-5). Note that the higher these metrics, the better the method.} 
\vspace{-5pt}
\begin{tabular}{rcccc}
\toprule
          & Alignment      & User-Q1         &  User-Q2          & User-Q3\\  
\midrule 
Text2Mesh~\cite{michel2022text2mesh}   & 0.262          & 1.750           &  1.506            & 1.472\\
TANGO~\cite{lei2022tango}       & 0.274          & 2.406           &  2.450            & 2.539\\
X-Mesh~\cite{ma2023x}      & 0.265          & 1.839           &  1.722            & 1.761\\
\midrule 
Our \ourMthd{}  & \textbf{0.285} & \textbf{3.344}  &  \textbf{3.311}   & \textbf{3.261}\\
\bottomrule
\end{tabular}
\vspace{-5pt}
\label{tab:quan}
\end{table}

\myPara{User Study.}
We further conduct a user study to evaluate these 3D stylization methods subjectively.
We randomly select 10 mesh-text pairs and recruit 60 users to evaluate the quality of the stylization assets generated by our \ourMthd{} and previous methods.
Particularly, the participants include experts in the field and individuals without specific background knowledge.
Moreover, each of them will be asked three questions~\cite{michel2022text2mesh}: (Q1) “How natural is the output results?”(Q2) “How well does the output match the original content?” (Q3) “How well does the output match the target style?”, and then assign a score (1-5) to them.
We report the mean opinion scores in parentheses for each factor averaged across all style outputs.
As shown in \tabref{tab:quan}, our \ourMthd{} still outperforms other methods across all questions.
Therefore, the 3D assets generated by our method are more in line with people's understanding of the text prompts.

\subsection{Ablation Studies} \label{subsec:abl}
To verify the effectiveness of the proposed designs in our \ourMthd{}, we conduct ablation studies by gradually adding them to our baseline model, \ie TANGO~\cite{lei2022tango}.
We chose the two-dragon mesh with the text prompt ``a fire dragon and an ice dragon'', and the experimental results are shown in \figref{fig:abl}.
Compared to the baseline model, introducing our \nameofmodule{} module enables the model to distinguish two dragons, yet it falls short in endowing them with precise texture details.
Meanwhile, incorporating our \nameofloss{} loss facilitates the model to capture more semantic details, \eg ``fire" and ``ice''.
Nevertheless, it fails to distinguish the two objects.
It is noteworthy that the model equipped with these two designs together is not only capable of accurately distinguishing between two objects but can also synthesize high-quality texture details for them.
These experiments indicate that our \nameofmodule{} module and \nameofloss{} loss can effectively assist the model in generating desired stylized content for multiple 3D objects to conform to the target text.

\section{Limitation and Future Work}
Despite achieving excellent results on text-driven multi-object stylization, our \ourMthd{} framework still has a few limitations, which can also facilitate future research:



\myPara{1) 3D Symmetry Prior.}
As stated in \secref{subsec:qual}, our \ourMthd{} fails to incorporate 3D symmetry prior, whose important role has been demonstrated by Text2Mesh~\cite{michel2022text2mesh} in promoting style consistency of a single object.
To generate more photorealistic stylization assets for multi-object scenes, it will be valuable to calculate symmetry planes for each object and apply symmetry priors to them.

\myPara{2) Diffusion Model.}
We observe that current diffusion technologies struggle to generate multi-object images according to the text prompt, which hinders the application of diffusion-based stylization methods in multi-object 3D scenes.
We argue it would be interesting to extend the concept of scene parsing to the diffusion models for the release of their potential in multi-object editing or generation.

\section{Conclusion}
In this paper, we present \ourMthd{}, an innovative framework proposing scene parsing and multi-grained cross-modal supervision to achieve text-driven multi-object 3D stylization for the first time.
%
%
Specifically, we first develop a \nameofmodule{} module to precisely align the objects in the 3D mesh and the text prompt and enhance the 3D point features with the word features belonging to the same object as them.
%
Then, we design a \nameofloss{} loss, in which the fine-grained loss at the local level and coarse-grained contrast loss at the global level are both constructed and complement each other.
Further, extensive experiments are conducted to demonstrate the effectiveness and superiority of our methods over the existing methods among a wide range of multi-object 3D scenes.
We believe it is promising to achieve content editing of multiple objects in 3D scenes simultaneously, and we hope the scene-parsing perspective provided by the proposed \ourMthd{} framework will inspire future works.
\clearpage
{
    \small
    \bibliographystyle{ieeenat_fullname}
    \bibliography{main}

\begin{thebibliography}{50}
\providecommand{\natexlab}[1]{#1}
\providecommand{\url}[1]{\texttt{#1}}
\expandafter\ifx\csname urlstyle\endcsname\relax
  \providecommand{\doi}[1]{doi: #1}\else
  \providecommand{\doi}{doi: \begingroup \urlstyle{rm}\Url}\fi

\bibitem[Bahdanau et~al.(2014)Bahdanau, Cho, and Bengio]{bahdanau2014neural}
Dzmitry Bahdanau, Kyunghyun Cho, and Yoshua Bengio.
\newblock Neural machine translation by jointly learning to align and translate.
\newblock \emph{arXiv preprint arXiv:1409.0473}, 2014.

\bibitem[Bird and Klein(2009)]{bird2009natural}
Edward~Loper Bird, Steven and Ewan Klein.
\newblock Natural language processing with python.
\newblock \emph{O'Reilly Media Inc}, 2009.

\bibitem[Boss et~al.(2021)Boss, Braun, Jampani, Barron, Liu, and Lensch]{boss2021nerd}
Mark Boss, Raphael Braun, Varun Jampani, Jonathan~T Barron, Ce Liu, and Hendrik Lensch.
\newblock Nerd: Neural reflectance decomposition from image collections.
\newblock In \emph{ICCV}, pages 12684--12694, 2021.

\bibitem[Caetano and Sra(2022)]{caetano2022arfy}
Arthur Caetano and Misha Sra.
\newblock Arfy: A pipeline for adapting 3d scenes to augmented reality.
\newblock In \emph{Adjunct Proceedings of the 35th Annual ACM Symposium on User Interface Software and Technology}, pages 1--3, 2022.

\bibitem[Chang et~al.(2015)Chang, Funkhouser, Guibas, Hanrahan, Huang, Li, Savarese, Savva, Song, Su, et~al.]{chang2015shapenet}
Angel~X Chang, Thomas Funkhouser, Leonidas Guibas, Pat Hanrahan, Qixing Huang, Zimo Li, Silvio Savarese, Manolis Savva, Shuran Song, Hao Su, et~al.
\newblock Shapenet: An information-rich 3d model repository.
\newblock \emph{arXiv preprint arXiv:1512.03012}, 2015.

\bibitem[Chen et~al.(2023{\natexlab{a}})Chen, Siddiqui, Lee, Tulyakov, and Nie{\ss}ner]{chen2023text2tex}
Dave~Zhenyu Chen, Yawar Siddiqui, Hsin-Ying Lee, Sergey Tulyakov, and Matthias Nie{\ss}ner.
\newblock Text2tex: Text-driven texture synthesis via diffusion models.
\newblock \emph{arXiv preprint arXiv:2303.11396}, 2023{\natexlab{a}}.

\bibitem[Chen et~al.(2023{\natexlab{b}})Chen, Chen, Jiao, and Jia]{chen2023fantasia3d}
Rui Chen, Yongwei Chen, Ningxin Jiao, and Kui Jia.
\newblock Fantasia3d: Disentangling geometry and appearance for high-quality text-to-3d content creation.
\newblock \emph{arXiv preprint arXiv:2303.13873}, 2023{\natexlab{b}}.

\bibitem[Chen et~al.(2022)Chen, Duinkharjav, Sun, Wei, Petrangeli, Echevarria, Silva, and Sun]{chen2022instant}
Shaoyu Chen, Budmonde Duinkharjav, Xin Sun, Li-Yi Wei, Stefano Petrangeli, Jose Echevarria, Claudio Silva, and Qi Sun.
\newblock Instant reality: Gaze-contingent perceptual optimization for 3d virtual reality streaming.
\newblock \emph{IEEE TVCG}, 28\penalty0 (5):\penalty0 2157--2167, 2022.

\bibitem[Devlin et~al.(2018)Devlin, Chang, Lee, and Toutanova]{devlin2018bert}
Jacob Devlin, Ming-Wei Chang, Kenton Lee, and Kristina Toutanova.
\newblock Bert: Pre-training of deep bidirectional transformers for language understanding.
\newblock \emph{arXiv preprint arXiv:1810.04805}, 2018.

\bibitem[Dosovitskiy et~al.(2020)Dosovitskiy, Beyer, Kolesnikov, Weissenborn, Zhai, Unterthiner, Dehghani, Minderer, Heigold, Gelly, et~al.]{dosovitskiy2020image}
Alexey Dosovitskiy, Lucas Beyer, Alexander Kolesnikov, Dirk Weissenborn, Xiaohua Zhai, Thomas Unterthiner, Mostafa Dehghani, Matthias Minderer, Georg Heigold, Sylvain Gelly, et~al.
\newblock An image is worth 16x16 words: Transformers for image recognition at scale.
\newblock \emph{arXiv preprint arXiv:2010.11929}, 2020.

\bibitem[Frans et~al.(2022)Frans, Soros, and Witkowski]{frans2022clipdraw}
Kevin Frans, Lisa Soros, and Olaf Witkowski.
\newblock Clipdraw: Exploring text-to-drawing synthesis through language-image encoders.
\newblock \emph{Advances in Neural Information Processing Systems}, 35:\penalty0 5207--5218, 2022.

\bibitem[Han et~al.(2022)Han, Guo, Chen, Liang, Zhao, Zhang, Bai, Zhang, Wei, Wu, et~al.]{han2022submillimeter}
Mengdi Han, Xiaogang Guo, Xuexian Chen, Cunman Liang, Hangbo Zhao, Qihui Zhang, Wubin Bai, Fan Zhang, Heming Wei, Changsheng Wu, et~al.
\newblock Submillimeter-scale multimaterial terrestrial robots.
\newblock \emph{Science Robotics}, 7\penalty0 (66):\penalty0 eabn0602, 2022.

\bibitem[Hu et~al.(2018)Hu, Shen, and Sun]{hu2018squeeze}
Jie Hu, Li Shen, and Gang Sun.
\newblock Squeeze-and-excitation networks.
\newblock In \emph{CVPR}, pages 7132--7141, 2018.

\bibitem[Lei et~al.(2022)Lei, Zhang, Jia, et~al.]{lei2022tango}
Jiabao Lei, Yabin Zhang, Kui Jia, et~al.
\newblock Tango: Text-driven photorealistic and robust 3d stylization via lighting decomposition.
\newblock \emph{NeurIPS}, 35:\penalty0 30923--30936, 2022.

\bibitem[Liu et~al.(2021)Liu, Lin, Cao, Hu, Wei, Zhang, Lin, and Guo]{liu2021swin}
Ze Liu, Yutong Lin, Yue Cao, Han Hu, Yixuan Wei, Zheng Zhang, Stephen Lin, and Baining Guo.
\newblock Swin transformer: Hierarchical vision transformer using shifted windows.
\newblock In \emph{ICCV}, pages 10012--10022, 2021.

\bibitem[Loshchilov and Hutter(2017)]{loshchilov2017decoupled}
Ilya Loshchilov and Frank Hutter.
\newblock Decoupled weight decay regularization.
\newblock \emph{arXiv preprint arXiv:1711.05101}, 2017.

\bibitem[Luo et~al.(2021)Luo, Ji, Sun, Cao, Wu, Huang, Lin, and Ji]{luo2021dual}
Yunpeng Luo, Jiayi Ji, Xiaoshuai Sun, Liujuan Cao, Yongjian Wu, Feiyue Huang, Chia-Wen Lin, and Rongrong Ji.
\newblock Dual-level collaborative transformer for image captioning.
\newblock In \emph{AAAI}, pages 2286--2293, 2021.

\bibitem[Luong et~al.(2015)Luong, Pham, and Manning]{luong2015effective}
Minh-Thang Luong, Hieu Pham, and Christopher~D Manning.
\newblock Effective approaches to attention-based neural machine translation.
\newblock \emph{arXiv preprint arXiv:1508.04025}, 2015.

\bibitem[Ma et~al.(2022)Ma, Xu, Sun, Yan, Zhang, and Ji]{ma2022x}
Yiwei Ma, Guohai Xu, Xiaoshuai Sun, Ming Yan, Ji Zhang, and Rongrong Ji.
\newblock X-clip: End-to-end multi-grained contrastive learning for video-text retrieval.
\newblock In \emph{ACM MM}, pages 638--647, 2022.

\bibitem[Ma et~al.(2023)Ma, Zhang, Sun, Ji, Wang, Jiang, Zhuang, and Ji]{ma2023x}
Yiwei Ma, Xiaioqing Zhang, Xiaoshuai Sun, Jiayi Ji, Haowei Wang, Guannan Jiang, Weilin Zhuang, and Rongrong Ji.
\newblock X-mesh: Towards fast and accurate text-driven 3d stylization via dynamic textual guidance.
\newblock \emph{arXiv preprint arXiv:2303.15764}, 2023.

\bibitem[Metzer et~al.(2023)Metzer, Richardson, Patashnik, Giryes, and Cohen-Or]{metzer2023latent}
Gal Metzer, Elad Richardson, Or Patashnik, Raja Giryes, and Daniel Cohen-Or.
\newblock Latent-nerf for shape-guided generation of 3d shapes and textures.
\newblock In \emph{CVPR}, pages 12663--12673, 2023.

\bibitem[Michel et~al.(2022)Michel, Bar-On, Liu, Benaim, and Hanocka]{michel2022text2mesh}
Oscar Michel, Roi Bar-On, Richard Liu, Sagie Benaim, and Rana Hanocka.
\newblock Text2mesh: Text-driven neural stylization for meshes.
\newblock In \emph{CVPR}, pages 13492--13502, 2022.

\bibitem[Mohammad~Khalid et~al.(2022)Mohammad~Khalid, Xie, Belilovsky, and Popa]{mohammad2022clip}
Nasir Mohammad~Khalid, Tianhao Xie, Eugene Belilovsky, and Tiberiu Popa.
\newblock Clip-mesh: Generating textured meshes from text using pretrained image-text models.
\newblock In \emph{SIGGRAPH Asia}, pages 1--8, 2022.

\bibitem[Poole et~al.(2022)Poole, Jain, Barron, and Mildenhall]{poole2022dreamfusion}
Ben Poole, Ajay Jain, Jonathan~T Barron, and Ben Mildenhall.
\newblock Dreamfusion: Text-to-3d using 2d diffusion.
\newblock \emph{arXiv preprint arXiv:2209.14988}, 2022.

\bibitem[Radford et~al.(2021)Radford, Kim, Hallacy, Ramesh, Goh, Agarwal, Sastry, Askell, Mishkin, Clark, et~al.]{radford2021learning}
Alec Radford, Jong~Wook Kim, Chris Hallacy, Aditya Ramesh, Gabriel Goh, Sandhini Agarwal, Girish Sastry, Amanda Askell, Pamela Mishkin, Jack Clark, et~al.
\newblock Learning transferable visual models from natural language supervision.
\newblock In \emph{ICML}, pages 8748--8763. PMLR, 2021.

\bibitem[Ramesh et~al.(2022)Ramesh, Dhariwal, Nichol, Chu, and Chen]{ramesh2022hierarchical}
Aditya Ramesh, Prafulla Dhariwal, Alex Nichol, Casey Chu, and Mark Chen.
\newblock Hierarchical text-conditional image generation with clip latents.
\newblock \emph{arXiv preprint arXiv:2204.06125}, 1\penalty0 (2):\penalty0 3, 2022.

\bibitem[Richardson et~al.(2023)Richardson, Metzer, Alaluf, Giryes, and Cohen-Or]{richardson2023texture}
Elad Richardson, Gal Metzer, Yuval Alaluf, Raja Giryes, and Daniel Cohen-Or.
\newblock Texture: Text-guided texturing of 3d shapes.
\newblock \emph{arXiv preprint arXiv:2302.01721}, 2023.

\bibitem[Rombach et~al.(2022)Rombach, Blattmann, Lorenz, Esser, and Ommer]{rombach2022high}
Robin Rombach, Andreas Blattmann, Dominik Lorenz, Patrick Esser, and Bj{\"o}rn Ommer.
\newblock High-resolution image synthesis with latent diffusion models.
\newblock In \emph{CVPR}, pages 10684--10695, 2022.

\bibitem[Roth(1982)]{roth1982ray}
Scott~D Roth.
\newblock Ray casting for modeling solids.
\newblock \emph{Computer graphics and image processing}, 18\penalty0 (2):\penalty0 109--144, 1982.

\bibitem[Sidi et~al.(2011)Sidi, Van~Kaick, Kleiman, Zhang, and Cohen-Or]{sidi2011unsupervised}
Oana Sidi, Oliver Van~Kaick, Yanir Kleiman, Hao Zhang, and Daniel Cohen-Or.
\newblock Unsupervised co-segmentation of a set of shapes via descriptor-space spectral clustering.
\newblock In \emph{SIGGRAPH Asia}, pages 1--10, 2011.

\bibitem[Tancik et~al.(2020)Tancik, Srinivasan, Mildenhall, Fridovich-Keil, Raghavan, Singhal, Ramamoorthi, Barron, and Ng]{tancik2020fourier}
Matthew Tancik, Pratul Srinivasan, Ben Mildenhall, Sara Fridovich-Keil, Nithin Raghavan, Utkarsh Singhal, Ravi Ramamoorthi, Jonathan Barron, and Ren Ng.
\newblock Fourier features let networks learn high frequency functions in low dimensional domains.
\newblock \emph{NeurIPS}, 33:\penalty0 7537--7547, 2020.

\bibitem[TurboSquid(2021)]{turbosquid2021turbosquid}
TurboSquid.
\newblock Turbosquid 3d model repository.
\newblock In \emph{https://www.turbosquid.com/}, 2021.

\bibitem[Vaswani et~al.(2017)Vaswani, Shazeer, Parmar, Uszkoreit, Jones, Gomez, Kaiser, and Polosukhin]{vaswani2017attention}
Ashish Vaswani, Noam Shazeer, Niki Parmar, Jakob Uszkoreit, Llion Jones, Aidan~N Gomez, {\L}ukasz Kaiser, and Illia Polosukhin.
\newblock Attention is all you need.
\newblock \emph{NeurIPS}, 30, 2017.

\bibitem[Wang et~al.(2022)Wang, Chai, He, Chen, and Liao]{wang2022clip}
Can Wang, Menglei Chai, Mingming He, Dongdong Chen, and Jing Liao.
\newblock Clip-nerf: Text-and-image driven manipulation of neural radiance fields.
\newblock In \emph{CVPR}, pages 3835--3844, 2022.

\bibitem[Wang et~al.(2018)Wang, Girshick, Gupta, and He]{wang2018non}
Xiaolong Wang, Ross Girshick, Abhinav Gupta, and Kaiming He.
\newblock Non-local neural networks.
\newblock In \emph{CVPR}, pages 7794--7803, 2018.

\bibitem[Wu et~al.(2022)Wu, Zhang, Sun, Zhou, Chen, Gu, Sun, and Ji]{wu2022difnet}
Mingrui Wu, Xuying Zhang, Xiaoshuai Sun, Yiyi Zhou, Chao Chen, Jiaxin Gu, Xing Sun, and Rongrong Ji.
\newblock Difnet: Boosting visual information flow for image captioning.
\newblock In \emph{CVPR}, pages 18020--18029, 2022.

\bibitem[Wu et~al.(2015)Wu, Song, Khosla, Yu, Zhang, Tang, and Xiao]{wu20153d}
Zhirong Wu, Shuran Song, Aditya Khosla, Fisher Yu, Linguang Zhang, Xiaoou Tang, and Jianxiong Xiao.
\newblock 3d shapenets: A deep representation for volumetric shapes.
\newblock In \emph{CVPR}, pages 1912--1920, 2015.

\bibitem[Yang et~al.(2021)Yang, Bisk, and Gao]{yang2021taco}
Jianwei Yang, Yonatan Bisk, and Jianfeng Gao.
\newblock Taco: Token-aware cascade contrastive learning for video-text alignment.
\newblock In \emph{ICCV}, pages 11562--11572, 2021.

\bibitem[Yao et~al.(2021)Yao, Huang, Hou, Lu, Niu, Xu, Liang, Li, Jiang, and Xu]{yao2021filip}
Lewei Yao, Runhui Huang, Lu Hou, Guansong Lu, Minzhe Niu, Hang Xu, Xiaodan Liang, Zhenguo Li, Xin Jiang, and Chunjing Xu.
\newblock Filip: Fine-grained interactive language-image pre-training.
\newblock \emph{arXiv preprint arXiv:2111.07783}, 2021.

\bibitem[Yin et~al.(2022)Yin, Zhang, Hou, Sun, Fan, and Van~Gool]{yin2022camoformer}
Bowen Yin, Xuying Zhang, Qibin Hou, Bo-Yuan Sun, Deng-Ping Fan, and Luc Van~Gool.
\newblock Camoformer: Masked separable attention for camouflaged object detection.
\newblock \emph{arXiv preprint arXiv:2212.06570}, 2022.

\bibitem[Yin et~al.(2023)Yin, Zhang, Li, Liu, Cheng, and Hou]{yin2023dformer}
Bowen Yin, Xuying Zhang, Zhongyu Li, Li Liu, Ming-Ming Cheng, and Qibin Hou.
\newblock Dformer: Rethinking rgbd representation learning for semantic segmentation.
\newblock \emph{arXiv preprint arXiv:2309.09668}, 2023.

\bibitem[Yin et~al.(2021)Yin, Gao, Shugrina, Khamis, and Fidler]{yin20213dstylenet}
Kangxue Yin, Jun Gao, Maria Shugrina, Sameh Khamis, and Sanja Fidler.
\newblock 3dstylenet: Creating 3d shapes with geometric and texture style variations.
\newblock In \emph{ICCV}, pages 12456--12465, 2021.

\bibitem[Zhang et~al.(2022{\natexlab{a}})Zhang, Goodman, and Gu]{zhang2022novel}
Bo Zhang, Lizbeth Goodman, and Xiaoqing Gu.
\newblock Novel 3d contextual interactive games on a gamified virtual environment support cultural learning through collaboration among intercultural students.
\newblock \emph{SAGE Open}, 12\penalty0 (2):\penalty0 21582440221096141, 2022{\natexlab{a}}.

\bibitem[Zhang et~al.(2021{\natexlab{a}})Zhang, Luan, Wang, Bala, and Snavely]{zhang2021physg}
Kai Zhang, Fujun Luan, Qianqian Wang, Kavita Bala, and Noah Snavely.
\newblock Physg: Inverse rendering with spherical gaussians for physics-based material editing and relighting.
\newblock In \emph{CVPR}, pages 5453--5462, 2021{\natexlab{a}}.

\bibitem[Zhang et~al.(2022{\natexlab{b}})Zhang, Kolkin, Bi, Luan, Xu, Shechtman, and Snavely]{zhang2022arf}
Kai Zhang, Nick Kolkin, Sai Bi, Fujun Luan, Zexiang Xu, Eli Shechtman, and Noah Snavely.
\newblock Arf: Artistic radiance fields.
\newblock In \emph{ECCV}, pages 717--733. Springer, 2022{\natexlab{b}}.

\bibitem[Zhang et~al.(2021{\natexlab{b}})Zhang, Srinivasan, Deng, Debevec, Freeman, and Barron]{zhang2021nerfactor}
Xiuming Zhang, Pratul~P Srinivasan, Boyang Deng, Paul Debevec, William~T Freeman, and Jonathan~T Barron.
\newblock Nerfactor: Neural factorization of shape and reflectance under an unknown illumination.
\newblock \emph{ACM TOG}, 40\penalty0 (6):\penalty0 1--18, 2021{\natexlab{b}}.

\bibitem[Zhang et~al.(2021{\natexlab{c}})Zhang, Sun, Luo, Ji, Zhou, Wu, Huang, and Ji]{zhang2021rstnet}
Xuying Zhang, Xiaoshuai Sun, Yunpeng Luo, Jiayi Ji, Yiyi Zhou, Yongjian Wu, Feiyue Huang, and Rongrong Ji.
\newblock Rstnet: Captioning with adaptive attention on visual and non-visual words.
\newblock In \emph{CVPR}, pages 15465--15474, 2021{\natexlab{c}}.

\bibitem[Zhang et~al.(2023)Zhang, Yin, Lin, Hou, Fan, and Cheng]{zhang2023referring}
Xuying Zhang, Bowen Yin, Zheng Lin, Qibin Hou, Deng-Ping Fan, and Ming-Ming Cheng.
\newblock Referring camouflaged object detection.
\newblock \emph{arXiv preprint arXiv:2306.07532}, 2023.

\bibitem[Zhou and Jacobson(2016)]{zhou2016thingi10k}
Qingnan Zhou and Alec Jacobson.
\newblock Thingi10k: A dataset of 10,000 3d-printing models.
\newblock \emph{arXiv preprint arXiv:1605.04797}, 2016.

\bibitem[Zivkovic(2004)]{zivkovic2004improved}
Zoran Zivkovic.
\newblock Improved adaptive gaussian mixture model for background subtraction.
\newblock In \emph{ICPR}, pages 28--31. IEEE, 2004.

\end{thebibliography}
}
\clearpage
\setcounter{page}{1}
\maketitlesupplementary

\section{Neural Stylization and Controls}
\begin{figure}[t]
    \centering
    \subfloat[a fire dragon and an ice dragon]{
        \includegraphics[width=.98\linewidth]{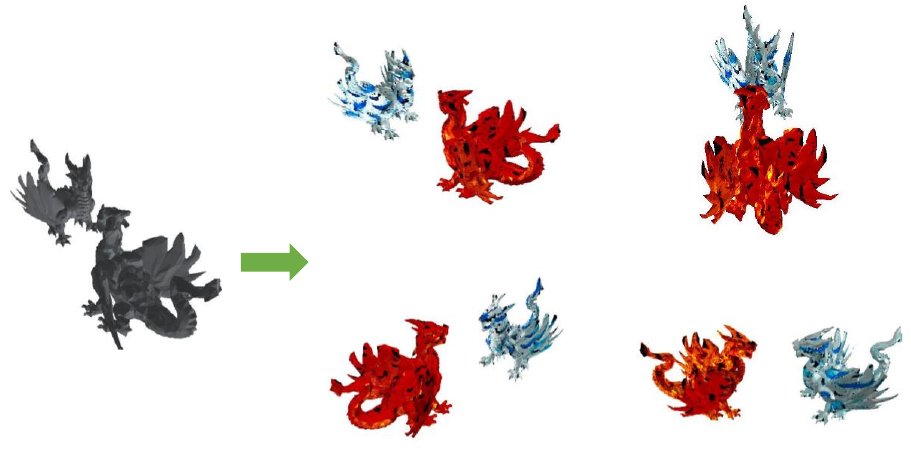}
    }\\
    \subfloat[a Garfield cat and a brown horse]{
        \includegraphics[width=.98\linewidth]{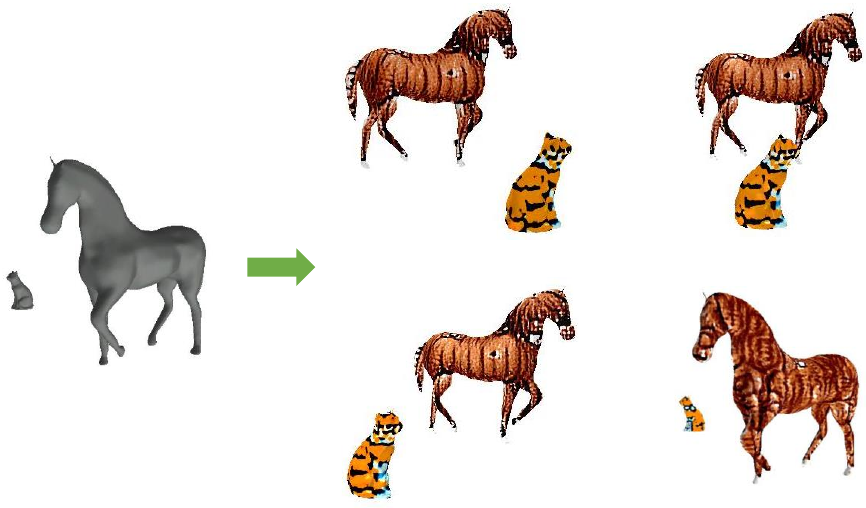}
    }\\
    \subfloat[a superman, an ice whale, and a fire dragon]{
        \includegraphics[width=.98\linewidth]{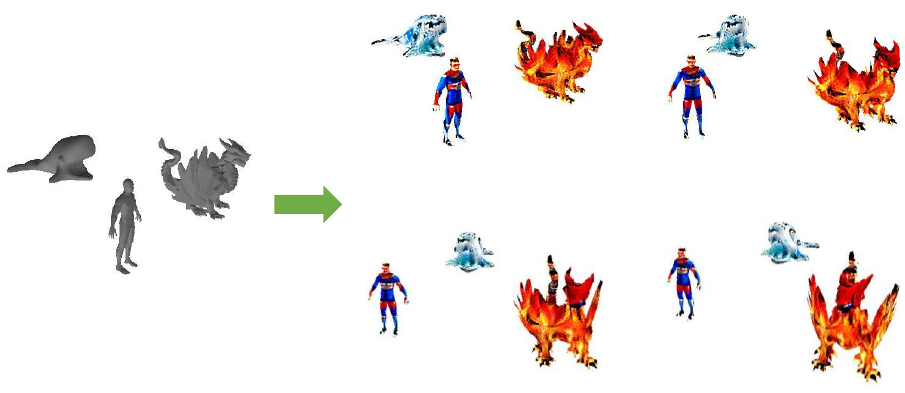}
    }
    \caption{Given several multi-object meshes, our \ourMthd{} stylizes entire 3D content on them to adhere to the text prompts. } 
  \label{fig:entire}
\end{figure}

\begin{figure*}[t]
    \centering
    \includegraphics[width=.98\linewidth]{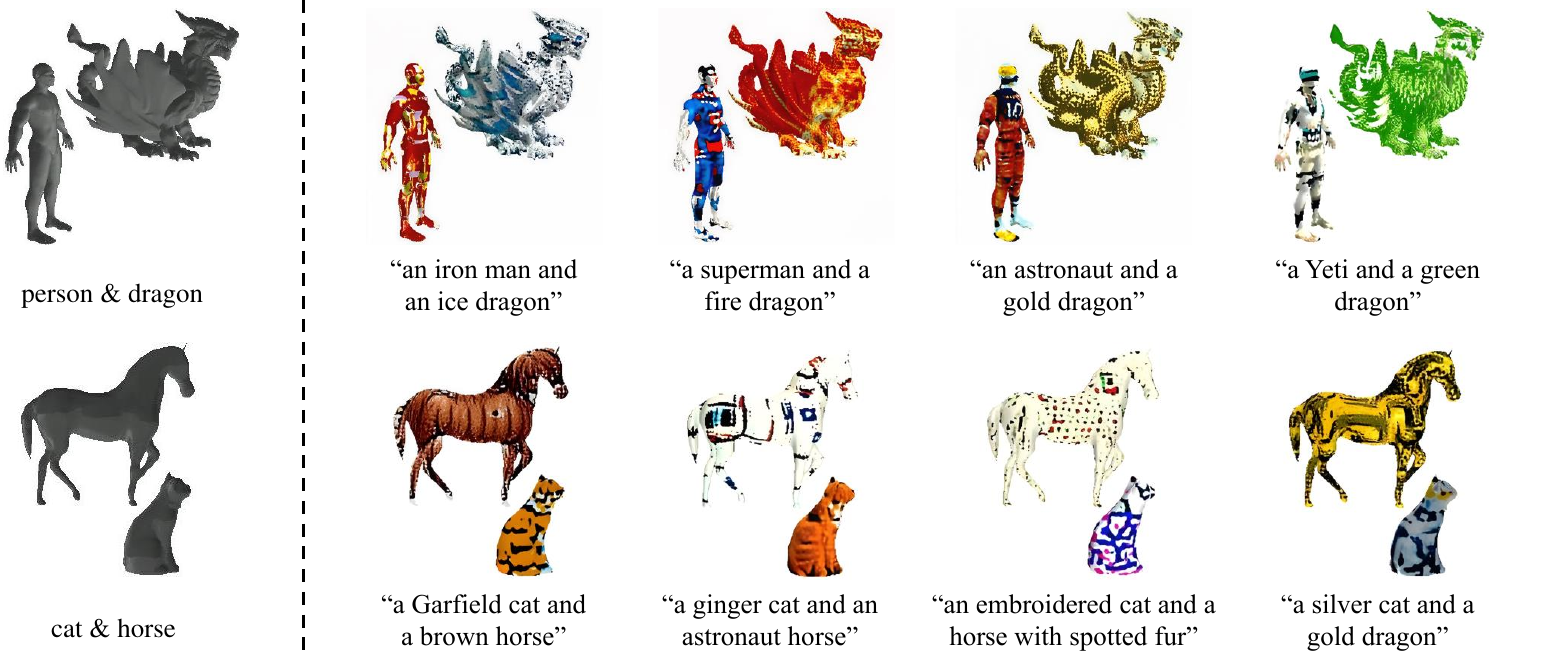}
    \caption{
    Given the same bare mesh, our \ourMthd{} method is able to produce stylized contents of high fidelity and various types for multi-object scenes to conform to the different text prompts.
    } 
    \vspace{8pt}
  \label{fig:disp_more}
\end{figure*}

\begin{figure*}[t]
    \centering
    \includegraphics[width=.98\linewidth]{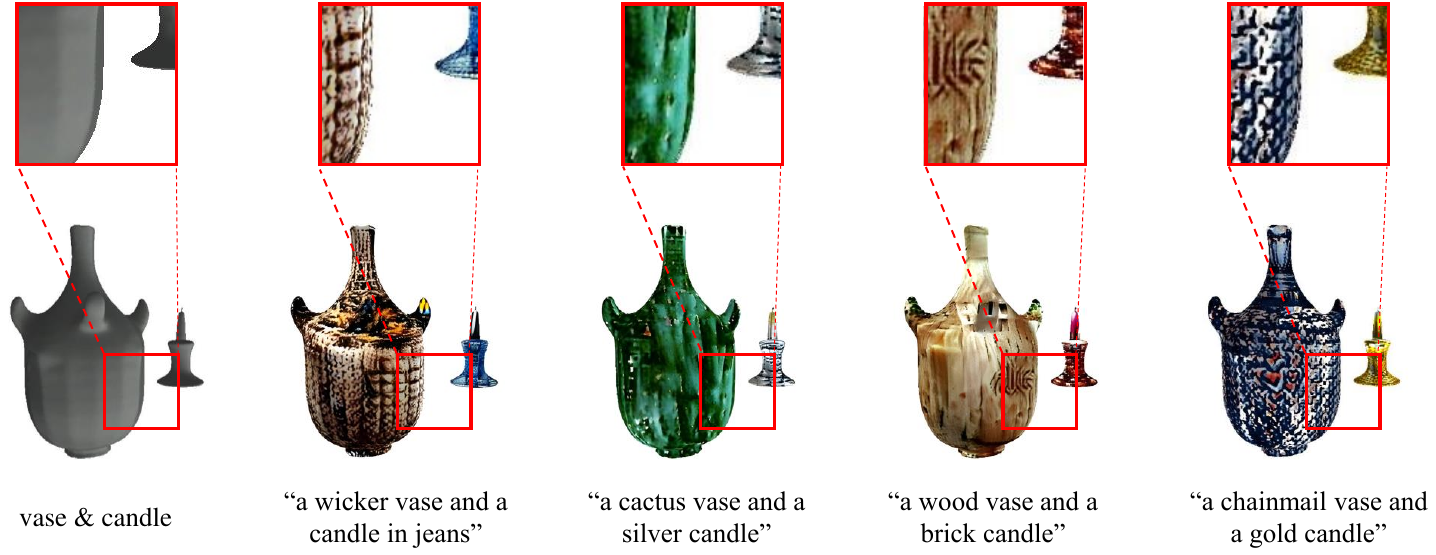}
    \caption{
    \ourMthd{} produces accurate and photorealistic details over a variety of multi-object scenes, driven by a series of text prompts.
    The local stylization results in red rectangle regions are zoomed in for better viewing.
    } 
  \label{fig:details}
\end{figure*}

In this section, we provide more details on the neural stylization and controls of our \ourMthd{} method.
We first render several stylized multi-object 3D assets of our \ourMthd{} from 4 views around them.
As shown in \figref{fig:entire}, the rendered images exhibit natural variation in texture and a high
degree of consistency across different viewpoints.
Take the cat-horse mesh coupled with a text prompt ``a Garfield cat and a brown horse'' as an example, our \ourMthd{} not only synthesize ``Garfield'' and ``brown'' property for the cat and horse separately but also generate visually plausible 3D content in different angles of view.
Then, we report more stylized results generated by our \ourMthd{}, in which each multi-object mesh is stylized according to several text prompts.
As shown in \figref{fig:disp_more}, our \ourMthd{} is able to synthesize stylized content faithful to different text prompts for the given mesh, which proves the robustness of our method for a variety of multi-object 3D scenes.
Furthermore, we also give details of the bare mesh and stylized 3D assets by zooming in their local regions.
As shown in \figref{fig:details}, our stylization results can capture both global semantics and part-aware details, conforming to the text prompts.
%
These experimental results indicate our \ourMthd{} is able to stylize the entire mesh in an object-consistent manner and flexibly generate results with accurate details as well as high fidelity.  
%


\begin{figure*}[t]
    \centering
    \footnotesize
    \begin{overpic}[width=0.98\linewidth]{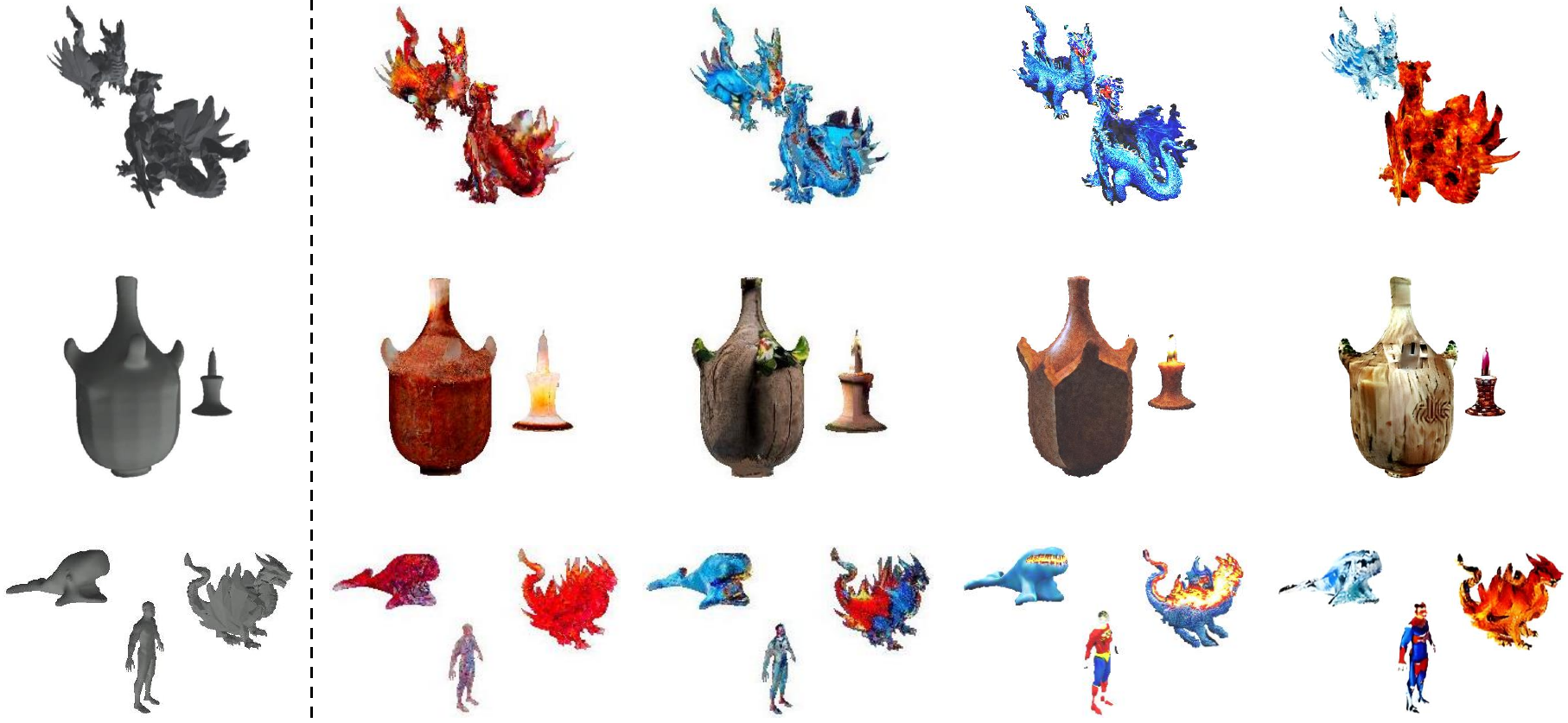}
    \put (5,46.6){\color{black}{\textbf{Bare Mesh}}}
    \put (26,46.6){\color{black}{\textbf{Latent-NeRF}}~\cite{metzer2023latent}}
    \put (45,46.6){\color{black}{\textbf{TEXTure}}~\cite{richardson2023texture}}
    \put (66,46.6){\color{black}{\textbf{Fantasia3D}}~\cite{chen2023fantasia3d}}
    \put (86,46.6){\color{black}{\textbf{Our \ourMthd{}}}}
    \put (5,30){two dragons}
    \put (44,30){Text Prompt: ``a fire dragon and an ice dragon''}
    \put (5,12.5){vase $\&$ candle}
    \put (44,12.5){Text Prompt: ``a wood vase and a brick candle''}
    \put (1,-2){person $\&$ dragon $\&$ whale}
    \put (41,-2){Text Prompt: ``a superman, a fire dragon, and an ice whale''}
    \end{overpic}
    \vspace{15pt}
    \caption{
    Visual comparisons of our \ourMthd{} with recent representative 3D stylization methods based on diffusion strategies on several multi-object scenes, including two objects of the same or different categories, and three different objects.
    } 
  \label{fig:comp2}
\end{figure*}

\begin{figure}[t]
    \centering
    \subfloat[a black cat and a white cat]{
        \includegraphics[width=.46\linewidth]{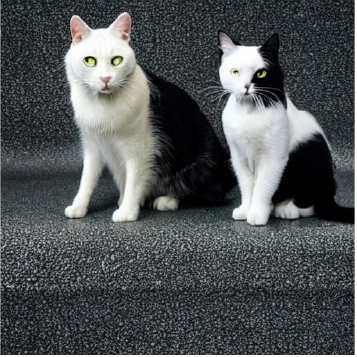}
    }
    \subfloat[a white cat and a brown dog]{
        \includegraphics[width=.46\linewidth]{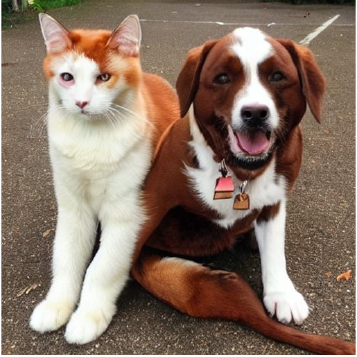}
    }
    \caption{Examples of multi-object 2D scenes generated by the cutting-edge text-to-image method, \ie stable-diffusion~\cite{rombach2022high}. For a scene with multiple objects of the same category or different categories, the diffusion model is prone to interference between different properties of the objects.} 
  \label{fig:sd_2d}
\end{figure}

\section{Qualitative Evaluation}
\myPara{Comparison with Diffusion-Based Methods.}
In this part, we compare the proposed \ourMthd{} with recent representative 3D stylization methods based on diffusion strategies, including Latent-NeRF~\cite{metzer2023latent} (CVPR~2023), TEXTure~\cite{richardson2023texture} (SIGGRAPH~2023), and Fantasia3D~\cite{chen2023fantasia3d} (ICCV~2023).
As shown in \figref{fig:comp2}, the existing diffusion-based methods are prone to interference between different properties of the objects for a scene with multiple objects of the same/different categories.
For the two-dragon mesh with a text prompt ``a fire dragon and an ice dragon'', these methods pay more attention to the ``fire'' or ``ice'' property in each object.
As far as the 3D scenes containing two or more different objects, these methods tend to focus on a certain property or mix multiple different properties together.
Differently, our TeMO equipped with 3D scene parsing and multi-grained supervision is able to accurately synthesize the desired stylized content for each object in all multi-object scenes.

\myPara{Analysis of Diffusion-based models on Multi-Object Scenes.}
Here, we try to analyze the reason why diffusion-based methods fail to stylize multi-object scenes without misunderstanding various properties.
Note that the existing diffusion-based methods utilize the priors in the pre-trained 2D text-to-image diffusion model by performing inference, on which basis the optimization of the 3D representation is achieved via a differential rendering.
We cannot help thinking about a straightforward question:
Could the pre-trained diffusion model generate accurate representations and images for the textual descriptions containing multiple objects?
Normally, the diffusion model employs the text encoder of the CLIP model~\cite{radford2021learning} to extract global semantic features of the text prompt for the guidance of image generation.
As discussed in \secref{sec:intro} of the main text, it is difficult for the CLIP model to encode the text description containing multiple objects with a global semantic representation.
We argue such an issue inevitably causes obstacles for diffusion methods in generating multi-object 2D scenes.
The poor multi-object results generated by the current cutting-edge diffusion methods for text-to-image like stable-diffusion~\cite{rombach2022high}, as shown in \figref{fig:sd_2d}, verify our hypothesis.
To address this issue, a foreseeable solution is to extend the concept of scene parsing proposed in this paper to the diffusion model.
We hope this perspective could inspire future works in the content editing of 2D~/3D multi-object scenes.


\end{document}